\documentclass{article}
\usepackage[T1]{fontenc} %
\usepackage{iclr2027_conference,times}

\usepackage{amsmath,amsfonts,bm}

\def\eqref#1{equation~\ref{#1}}

\def\1{\bm{1}}

\DeclareMathAlphabet{\mathsfit}{\encodingdefault}{\sfdefault}{m}{sl}
\SetMathAlphabet{\mathsfit}{bold}{\encodingdefault}{\sfdefault}{bx}{n}

\usepackage{booktabs}
\usepackage{graphicx}
\usepackage{hyperref}
\usepackage[percent]{overpic}
\usepackage{subcaption}
\usepackage{tabularx}
\usepackage{url}
\usepackage{wrapfig}
\usepackage{float}
\usepackage{placeins}
\usepackage{xcolor}

\definecolor{CatNTColor}{RGB}{19,131,139}
\definecolor{ContNTColor}{RGB}{196,72,43}
\definecolor{CatIterColor}{RGB}{165,54,128}
\definecolor{ContIterColor}{RGB}{67,102,178}

\DeclareRobustCommand{\CatNT}{%
  \texorpdfstring{\textcolor{CatNTColor}{\textsc{Cat-NT}}}{Cat-NT}}
\DeclareRobustCommand{\ContNT}{%
  \texorpdfstring{\textcolor{ContNTColor}{\textsc{Cont-NT}}}{Cont-NT}}
\DeclareRobustCommand{\CatIter}{%
  \texorpdfstring{\textcolor{CatIterColor}{\textsc{Cat-Iter}}}{Cat-Iter}}
\DeclareRobustCommand{\ContIter}{%
  \texorpdfstring{\textcolor{ContIterColor}{\textsc{Cont-Iter}}}{Cont-Iter}}

\title{Beyond Visual Fidelity: Comparing\\
Categorical and Continuous World Models}

\author{%
{Sarah Parisot, Melanie Fernandez Pradier, David Bignell,}\\
{\textbf{Sergio Valcarcel Macua, Katja Hofmann}}\\
{Microsoft}
}

\iclrfinalcopy
\begin{document}

\maketitle
\lhead{Preprint}

\begin{abstract}
Visual world models can now generate compelling interactive worlds, but visual realism does not guarantee faithful dynamics.
{We systematically compare how categorical and continuous prediction shape learned dynamics under both next-token and iterative whole-frame generation. These choices are often bundled in approaches such as autoregressive token models and whole-frame flow matching.}
{We design controlled 2D and 3D environments and evaluate dynamics validity, calibration, and visual quality.}
{In 2D with lossless visual encoding, categorical models achieve higher rollout validity with less training compute, particularly for stochastic entities.}
{Among categorical models, greedy decoding produces the highest rollout validity, while sampling from a next-token model yields the best-calibrated stochastic predictions. Failures in continuous models, such as disappearing game characters, echo those seen in more complex environments. Here, known dynamics let us measure their impact. In 3D, categorical models also achieve stronger long-horizon performance with less training compute. Even models with lossy tokenization remain competitive on semantic accuracy and FVD, despite lower next-frame RGB accuracy. Together, these findings make a strong case for categorical prediction as a compute-efficient route to learning world dynamics.}

\end{abstract}

\section{Introduction}

Recent advances in video generation have made visual world models, i.e., learned simulators of how environments evolve under actions, promising tools for planning, policy learning, and interactive applications \citep{UniPi,DIAMOND,WHAM,GameNGen}. However, many applications require more than visual realism: generation must respond faithfully to actions and preserve the underlying dynamics. How best to learn these dynamics remains unclear.

Approaches to learning these dynamics include autoregressive prediction of discrete visual tokens (categorical prediction), building on language-model architectures \citep{vaswani2017attention}, and continuous generation methods such as flow matching \citep{FlowMatching}. Continuous approaches are often preferred for visual generation because token-based models typically rely on lossy encodings that can limit visual quality. Research on narrowing this gap \citep{VisualAR,LlamaGen,MAGVITv2} shows that improvements in tokenization can change their relative performance. Yet visually compelling predictions can still violate physical interactions, change object state incorrectly, or lose consistency over time \citep{sora}. {\citet{Orbis}} directly compare categorical and continuous world models using a shared hybrid tokenizer, finding an advantage for continuous modeling in driving scenes. Yet tokenization-induced differences in visual fidelity make it difficult to attribute this advantage to dynamics modeling alone. We therefore ask: how do categorical and continuous prediction compare when visual encoding introduces no information loss?

\begin{figure}[t]
    \centering
    \includegraphics[width=\linewidth]{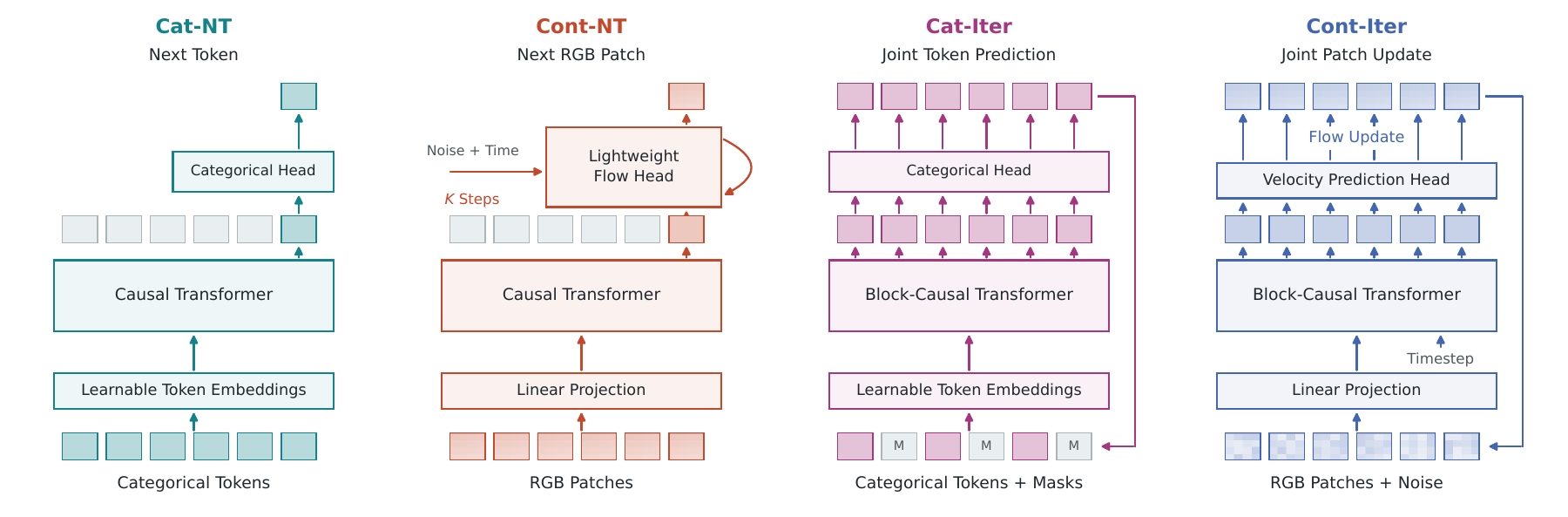}
    \caption{{\textbf{Architecture design.} We construct four architectures that systematically cross categorical and continuous prediction with next-token and iterative whole-frame generation.}}
    \label{fig:archi}
    \vspace{-0.5cm}
\end{figure}

We address this question through a controlled comparison that {crosses prediction representation with within-frame generation procedure}. Both categorical and continuous predictions can be generated sequentially, patch by patch~\citep{IRIS,li2024autoregressive}, or jointly through iterative updates~\citep{Genie,DIAMOND}. Flow matching, for example, can generate either individual patches or a complete frame. Our resulting four-model design crosses these choices (Figure~\ref{fig:archi}). 
{This allows us to compare the resulting combinations under shared environment and evaluation protocols.}
We study these choices in two complementary environments. A Pac-Man-inspired environment admits lossless encoding and lets us vary the number and behavior of interacting entities. Its known rules allow us to assess whether generated transitions and trajectories are permissible, distinguishing violations of the dynamics from legitimate stochastic alternatives without requiring agreement with a single reference. A partially observed 3D maze extends the comparison to examine how these findings change in 3D spaces and when learned lossy tokenization is introduced.

\paragraph{Contributions.} Our central result is that, with lossless visual encoding in Pac-Man, categorical world models {can} sustain valid trajectories at lower training compute than continuous models, with {the} largest differences in performance arising from stochastic dynamics.
These failures echo reports of disappearing or temporally inconsistent {non-playable} entities in large, high-fidelity diffusion and flow-based world models~{\citep{DIAMOND,Genie2}}, but our controlled setting makes their frequency and consequences measurable. The categorical advantage does not extend uniformly to reproducing outcome probabilities: continuous models can outperform iterative categorical prediction in calibration despite weaker validity. This distinguishes learning which futures are possible from learning how likely they are. The 3D maze experiments further separate reconstruction fidelity from rollout performance: lossy tokenization reduces next-frame visual accuracy without consistently degrading long-horizon semantic or video-level metrics. Decoding choices also matter, with sampling improving maze Fr\'{e}chet Video Distance (FVD) while greedy decoding favors Pac-Man validity. Together, these results show why representation and generation choices must be assessed against the properties required by the intended application, and how controlled environments complement large-scale evaluation {by directly measuring dynamics validity and calibration}.

\section{Related Work}

\noindent\textbf{Discrete and continuous world models.}
Discrete-token world models \citep{WHAM,MineWorld,Genie} demonstrate
controllable interactive generation using categorical sequence models. In
contrast, many recent frontier systems favor continuous diffusion-based
generation \citep{Matrix,Genie2,DIAMOND,GameNGen}, motivated in part by
improved visual fidelity and the avoidance of lossy quantization. Together, these works highlight a potential trade-off between visually faithful observations and stable, controllable dynamics over long rollouts.

\noindent\textbf{Comparing output parameterizations.}
{\citet{Orbis}} provide the closest direct comparison between continuous and discrete world models, finding that continuous flow matching outperforms discrete autoregression in visually complex driving scenes. 
Conversely, image-generation studies show that discrete models can match or
outperform diffusion with sufficiently strong tokenizers and architectures
\citep{LlamaGen,MAGVITv2,VisualAR}. 
These end-to-end comparisons couple output parameterization with tokenization, architecture, model scale, and compute, making their individual contributions difficult to distinguish. %
{Our four-model comparison systematically crosses output parameterization with generation procedure, revealing how these choices affect dynamics validity and calibration differently.}

\noindent\textbf{Long-horizon stochastic dynamics.} {Video-generation benchmarks such as VBench emphasize perceptual quality and consistency \citep{vbench}, while recent world-model evaluations explicitly go beyond visual fidelity. WorldModelBench \citep{WorldModelBench} evaluates instruction following and physics adherence, WorldScore \citep{WorldScore} measures controllability, quality, and dynamics, and World-in-World \citep{WorldInWorld} evaluates closed-loop embodied task success.}
{Long-horizon evaluation is particularly important because local errors can accumulate as models increasingly condition on their own predictions \citep{GameNGen,chen2025diffusion,huang2025selfforcing}. How this accumulation differs between categorical and continuous models as the number and stochasticity of interacting entities increase remains unclear. We complement existing work through controlled comparisons under lossless encoding and sequential or iterative generation, evaluating dynamics validity and calibration against known transition rules and probabilities.}

\section{World Model Architectures For a Controlled Comparison}
\label{sec:architectures}
\vspace{-0.2cm}

We compare four action-conditioned world models in a $2\times2$ design crossing output representation, namely categorical or continuous, with generation procedure, namely next-token prediction or iterative whole-frame generation (see Fig.~\ref{fig:archi} for a schematic overview).
All use Transformer backbones \citep{vaswani2017attention}. Next-token variants use pre-normalization decoder-only blocks \citep{GPT2,xiong2020layer}, while iterative models use DiT-style blocks \citep{DiT}.
We hold training data and optimization schedule fixed, and approximately match backbone capacity. Because the models require different input and output heads, enforcing identical total parameter counts would introduce additional architectural differences. We instead report performance as a function of cumulative training compute. %
Further implementation details are provided in App.~\ref{app:world_model_training}.

\noindent\textbf{Observation and action representation.}
For each environment, an observation $x_t$ is represented as a sequence of $P$ image patches. Categorical models encode each patch as a token $z_{t,p}\in\{1,\ldots,V\}$, where $V$ is the categorical vocabulary size. 
Continuous models represent each patch by a vector $y_{t,p}\in\mathbb{R}^{d}$, where $d$ is the continuous patch dimension and $p\in\{1,\ldots,P\}$. Within each environment, the two categorical models share the same observation encoding, as do the two continuous models. Environment-specific encodings are described in Sections~\ref{subsec:env-Pac-Man} and~\ref{subsec:env-mazeballs}.
All models predict $x_{t+1}$ conditioned on a context history of at most $L$ frames and actions $\mathcal{C}_t=(x_{t-L+1},a_{t-L+1},\ldots,x_t,a_t)$, where $a_t$ is the action leading from $x_t$ to the target frame $x_{t+1}$. The context length $L$ is fixed within each experiment. All models use learned absolute positional embeddings over their flattened observation--action sequences and are trained to predict the target frame from a ground-truth context (teacher-forcing). 
At inference time, generated frames are fed back into the temporal context.

\subsection{Next-token World Models}
\label{sec:causal_models}

We call ``next-token models'' those action-conditioned autoregressive world models that represent environment dynamics as next-token prediction \citep{IRIS,WHAM}, {generating the next frame autoregressively, patch by patch}. Our two %
variants use the same sequentially-ordered factorization with either categorical or continuous patch distributions:
\begin{equation}
    p_\theta(x_{t+1}\mid\mathcal{C}_t)
=
    \prod_{p=1}^{P}
    p_\theta(x_{t+1,p}\mid\mathcal{C}_t,x_{t+1,<p})
,
\end{equation}
where $x_{t+1,<p}$ denotes the target patches preceding position $p$. Both variants use the same decoder-only transformer backbone. Context patches and actions are interleaved in a single sequence, and each generated patch is appended to this sequence before the next patch is predicted.

\textbf{Categorical next-token model.}
\CatNT{} instantiates the causal factorization with categorical patch
distributions. 
It embeds visual patches as categorical tokens. Discrete actions share the embedding table using a separate identifier range, whereas continuous actions are linearly projected into the hidden space.
The visual embeddings are tied to the output classifier, and the model minimizes cross-entropy over the target-frame tokens:
\begin{equation}
    \mathcal{L}_{\mathrm{CatNT}}(\theta)
=
    -
    \sum_{p=1}^P
    \log p_\theta
    \left(
        z_{t+1,p}
        \mid
        \mathcal{C}_t,
        z_{t+1,<p}
    \right)
.
\end{equation}
At inference, visual tokens are generated sequentially, either greedily or using categorical sampling.

\textbf{Continuous next-token model.}
To our knowledge, no prior world model combines causal, patch-by-patch next-frame prediction with a separate continuous generative process for each patch. 
We select this architecture to enable a controlled comparison between categorical and continuous heads.
\ContNT{} adapts per-token diffusion heads from autoregressive image and multimodal generation \citep{li2024autoregressive,NextTokenDiffusion} to the action-conditioned world-model setting, and replaces the diffusion objective with conditional flow matching. 
It encodes RGB patches into the transformer hidden space and uses learned
embeddings for discrete actions or linear projections for continuous actions.
For target position $p$, the transformer produces the conditioning vector $h_{t+1,p}=f_\theta(\mathcal{C}_t,y_{t+1,<p})$. A conditional flow head then generates the patch from Gaussian noise.

Following conditional flow matching \citep{FlowMatching}, let $\epsilon\sim\mathcal{N}(0,I)$ and 
$
    \tilde y_{\tau}
=
    (1-\sigma_\tau)
    y
    +
    \sigma_\tau\epsilon
$, 
where $\sigma_\tau$ specifies the flow schedule. 
The target velocity with respect to the noise level is $\epsilon-y$.
For \ContNT{}, the clean sample is a target patch $y=y_{t+1,p}$ and the conditioning variable is $h=h_{t+1,p}$:
\begin{equation}
\label{eq:conditional_flow_matching}
    \mathcal{L}_{\mathrm{ContNT}} (\theta, \phi)
    =
    \mathbb{E}_{\tau,\epsilon}
    \left[
        \left\|
            v_\phi
            (
                \tilde y_\tau,
                \tau\mid 
                f_\theta(\mathcal{C}_t,y_{t+1,<p})
            )
            -
            \left(
                \epsilon
                -
                y_{t+1,p}
            \right)
        \right\|_2^2
    \right].
\end{equation}
At inference, the velocity field is integrated from noise level one to zero separately for each patch.
The resulting patch is encoded into the transformer hidden space and appended to the sequence before the next patch is generated.
Thus, \CatNT{} and \ContNT{} share similar factorization and transformer backbone but differ in whether each conditional distribution is represented by categorical logits or a continuous flow.

\subsection{Iterative Whole-Frame World Models}
\label{sec:iterative_models}

Iterative whole-frame generation has been adopted in world models through masked categorical refinement {\citep{Genie,WHAMRT}} and continuous diffusion or flow-based generation {\citep{DIAMOND,GameNGen,Orbis}}. Rather than imposing an ordering within the target frame, they represent all target patches simultaneously and update them over multiple iterations.
Both our iterative model variants use DiT-style blocks with block-causal temporal attention: patches interact bidirectionally within a frame but attend only to the current and preceding temporal blocks. The variants differ in their input and output layers, conditioning mechanisms, and update rules.

\textbf{Categorical iterative model.}
\CatIter{} uses the MaskGIT training and decoding procedure \citep{MaskGIT}, which has been adapted to world modeling in different ways. Genie applies masked prediction within a spatiotemporal dynamics model conditioned on learned latent actions \citep{Genie}, whereas {WHAM-RT combines one-shot next-frame token prediction with lightweight iterative masked refinement conditioned on a compressed temporal context \citep{WHAMRT}}. Here, we use the same masked-token principle as {WHAM-RT while aligning backbone and conditioning for controlled comparison.}

During training, a random subset $\mathcal{M}\subseteq\{1,\ldots,P\}$ of target positions is replaced by a learned mask embedding. 
The model predicts the masked tokens conditioned on the context, action, and visible target tokens $z_{t+1,\overline{\mathcal M}}$, where
$\overline{\mathcal M}=\{1,\ldots,P\}\setminus\mathcal M$:
\begin{equation}
    \mathcal{L}_{\mathrm{CatIter}}(\theta)
=
    -\sum_{p\in\mathcal{M}}
    \log p_\theta
    \left(
        z_{t+1,p}
        \mid
        \mathcal{C}_t,
        z_{t+1,\overline{\mathcal{M}}}
    \right)
.
\end{equation}
At inference, all target positions are initially masked. At each iteration, the model predicts the remaining tokens and permanently reveals the highest-confidence predictions according to a linear schedule, continuing until the target frame is complete.

\textbf{Continuous iterative flow model.}
\ContIter{} instantiates the continuous iterative variant using a flow-matching objective. Continuous context patches and their corresponding actions are interleaved in the context sequence, and the noised target patches are appended as the newest temporal block. Flow-time embeddings modulate the DiT-style blocks and output layer through adaptive normalization.
It uses the same conditional flow-matching objective as Equation~\ref{eq:conditional_flow_matching}, but applies it jointly to the complete patchified target frame $y_{t+1}=(y_{t+1,1},\ldots,y_{t+1,P})$,
so
$
    \tilde y_{\tau}
=
    (1-\sigma_\tau)
    (y_{t+1,1},\ldots,y_{t+1,P})
    +
    \sigma_\tau\epsilon
$. 
Conditioned on $\mathcal{C}_t$, its backbone predicts a coupled velocity for all target patches rather than a separate conditional flow for each patch.
At inference, the complete target frame is initialized from Gaussian noise and all patches are updated jointly by integrating the learned velocity field. This contrasts with \ContNT{}, which integrates a separate conditional flow for each patch and feeds the generated patches back sequentially.

\begin{figure}[t]
    \centering
    \begin{subfigure}[t]{0.35\textwidth}
        \centering
        \vspace{6pt}
        \includegraphics[
            height=0.58\textwidth,
            keepaspectratio
        ]{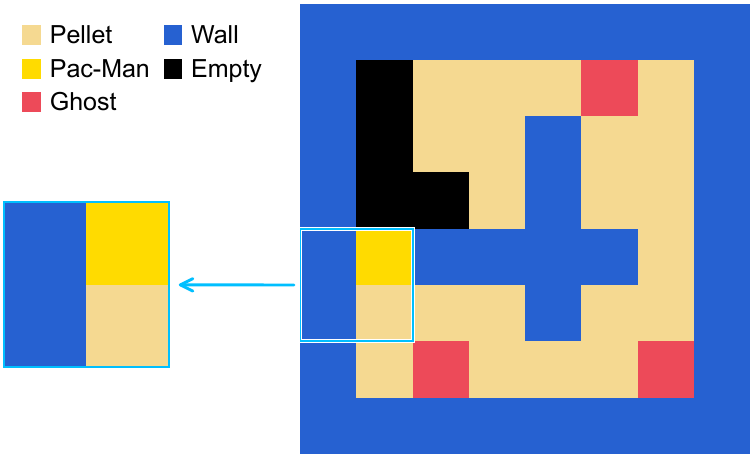}
        \caption{Pixel Pac-Man}
        \label{fig:Pac-Man}
    \end{subfigure}
    \hfill
    \begin{subfigure}[t]{0.64\textwidth}
        \centering
        \vspace{0pt}
        \includegraphics[
            height=0.34\textwidth,
            keepaspectratio
        ]{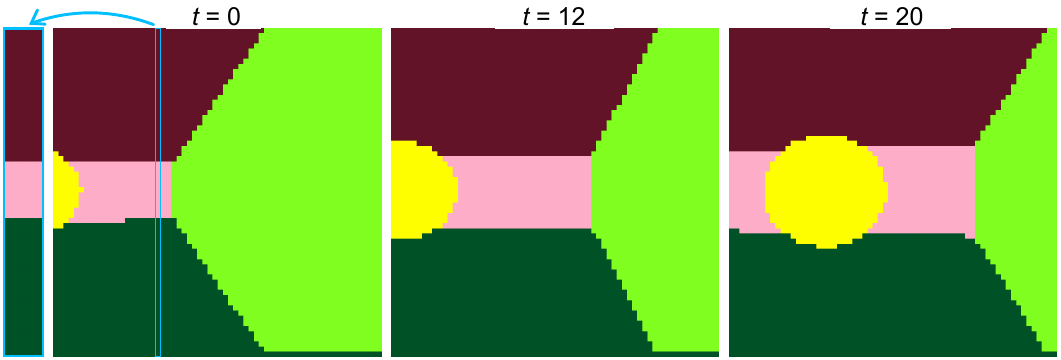}
        \caption{3D maze}
        \label{fig:maze}
    \end{subfigure}

    \caption{{\textbf{Environments designed for the controlled comparisons in this paper: Pixel Pac-Man and 3D Maze.}
    Cyan outlines and arrows indicate extracted patches: a
    $2 \times 2$ tile patch in Pac-Man and a $64 \times 1$ pixel
    column in the maze.}} %
    \label{fig:environments}
    \vspace{-0.4cm}
\end{figure}

\vspace{-0.2cm}
\section{Experiments}
\label{sec:experiments}
\vspace{-0.2cm}
In this section we describe our controlled toy environments and provide the main findings obtained with each of them.
{Further details are provided in Appendix~\ref{app:world_model_training}.}

\subsection{``Pixel Pac-Man'' Controlled Environment}
\label{subsec:env-Pac-Man}

Our first environment is a Pac-Man-inspired pixel-navigation task defined on an $N\times N$ grid. Walls and pellets occupy individual cells, as do the controllable agent and each of the $K$ ghosts. The agent aims to consume all pellets while avoiding collisions with ghosts. The primary comparison uses an $8\times8$ grid, three ghosts, and a fixed wall layout. Figure~\ref{fig:Pac-Man} shows an example.
The agent has five actions: up, down, left, right, and stay. Movement is
restricted to orthogonally adjacent non-wall cells, and actions directed into
walls leave the agent stationary. Moving onto a pellet consumes it. After the agent acts, ghosts are updated sequentially, avoiding cells occupied by other ghosts. The number of ghosts and their behaviour are configurable. 
We consider two ghost policies: uniform \emph{random} movement and \emph{flee}, which maximizes the Manhattan distance to the agent, breaking ties randomly. 
The primary comparison uses random ghosts that move at every step.
An episode terminates when the agent collides with a ghost or consumes the final pellet. Following termination, the wall layout is preserved, pellets are restored to all traversable cells except the agent's starting cell, and the agent and ghosts return to deterministic reset positions. Initial states used to generate training trajectories instead randomize the agent and ghost positions and sample different pellet densities. Training actions are sampled uniformly from the five-action space.

\noindent \textbf{Observation encoding.}
Each observation is a four-channel binary image indicating walls, pellets,
the agent, and ghosts. The underlying environment state admits seven valid
per-pixel occupancy patterns: empty, wall, pellet, agent, ghost, ghost with
pellet, and an agent-ghost collision. In the primary observation setting,
pellets covered by ghosts remain in the underlying state but are hidden from
the observed pellet channel.
We divide each $8\times8$ observation into sixteen non-overlapping
$2\times2$ patches, i.e., $P=16$. Categorical models encode each patch by
enumerating all combinations of the seven valid cell states across its four
pixels, giving a vocabulary of $V=7^4=2{,}401$ tokens. This encoding is
lossless for every valid observation and therefore introduces no tokenizer
reconstruction error.
For continuous models, we render each occupancy pattern using a distinct fixed
RGB colour. Each $2\times2$ RGB patch is flattened and rescaled to
$[-1,1]^{3\times2\times2}$, giving $d=12$. 
Continuous predictions are decoded by mapping each generated pixel to its
nearest palette entry. The projected frames are used for evaluation and
fed back as context during rollout.
All Pixel Pac-Man models condition on at most $L=15$ preceding frames.

\noindent \textbf{Evaluation metrics.}
The Pac-Man environment has stochastic dynamics, so disagreement with a sampled reference trajectory does not necessarily indicate a prediction error. A key advantage of our constructed environment is that its dynamics are known, allowing us to assess whether generated transitions are permissible rather than requiring agreement with a single reference trajectory. We evaluate this validity at two levels. \textbf{Transition validity} is the fraction of predicted transitions satisfying the relevant simulator constraints, irrespective of earlier failures. \textbf{Rollout validity} measures consistency over time: at each horizon, it is the fraction of trajectories that remain valid throughout their generated prefix. We evaluate 64-step autoregressive rollouts. Once a trajectory fails, subsequent valid transitions do not restore its rollout validity. Where appropriate, we further split transition and rollout validity by component (ghosts, Pac-Man, board), or into stochastic (ghosts) and deterministic (Pac-Man and board) elements.
Further details are provided in Appendix~\ref{app:pacman-metrics}.

\subsubsection{Findings in Pixel Pac-Man}
\label{subsec:pacman-findings}

\begin{figure}[t]
    \centering

    \includegraphics[width=\linewidth]{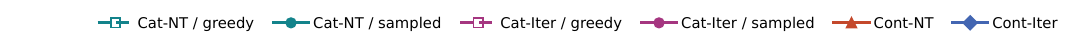}
    \vspace{-25pt}

\begin{subfigure}[t]{0.34\linewidth}
    \centering
    \phantomsubcaption
    \label{fig:next-frame-overall-validity}
    \begin{overpic}[width=\linewidth]
        {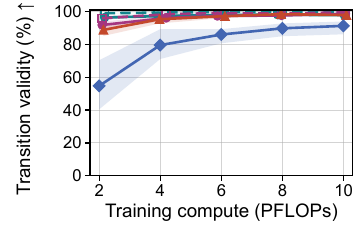}
        \put(84,24){\colorbox{white}{\textbf{(\thesubfigure)}}}
    \end{overpic}
\end{subfigure}
\hfill
\begin{subfigure}[t]{0.283\linewidth}
    \centering
    \phantomsubcaption
    \label{fig:next-frame-deterministic-validity}
    \begin{overpic}[width=\linewidth]
        {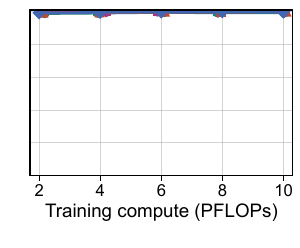}
        \put(80,29){\colorbox{white}{\textbf{(\thesubfigure)}}}
    \end{overpic}
\end{subfigure}
\hfill
\begin{subfigure}[t]{0.34\linewidth}
    \centering
    \phantomsubcaption
    \label{fig:next-frame-calibration}
    \begin{overpic}[width=\linewidth]{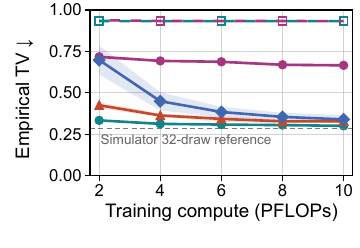}
        \put(85,24){\colorbox{white}{\textbf{(\thesubfigure)}}}
    \end{overpic}
\end{subfigure}

    \vspace{-0.4em}
    \caption{{
        \textbf{Validity and calibration across models.}
        \textbf{(a)} Overall next-frame transition validity.
        \textbf{(b)} Validity of deterministic components (Pac-Man and board).
        \textbf{(c)} Empirical total variation (TV) between predicted ghost configurations and the simulator's exact conditional distribution.
        The dashed line shows expected empirical TV.
        Unless specified otherwise, lines show means across training seeds and shading shows their minimum--maximum range throughout this paper.
    }}
    \label{fig:next_frame}
\end{figure}

\noindent \textbf{{Finding 1: All models achieve near-perfect next-frame validity on deterministic components.}} We first compare next-frame transition validity across training-compute budgets (Figure~\ref{fig:next_frame}(a)).
Overall validity improves with training compute for all approaches. The categorical models perform well even at low compute, whereas the continuous models, especially \ContIter{}, show larger improvements with training but remain less reliable. {To distinguish learning deterministic dynamics from modeling stochastic outcomes, Figure~\ref{fig:next_frame}(b) evaluates the deterministic components (Pac-Man and board) without ghost-dynamics checks. All approaches achieve near-perfect validity on these components from 2 PFLOPs, with the main differences arising from stochastic ghost dynamics.}

\begin{figure}[t]
    \centering
    \includegraphics[width=\linewidth]{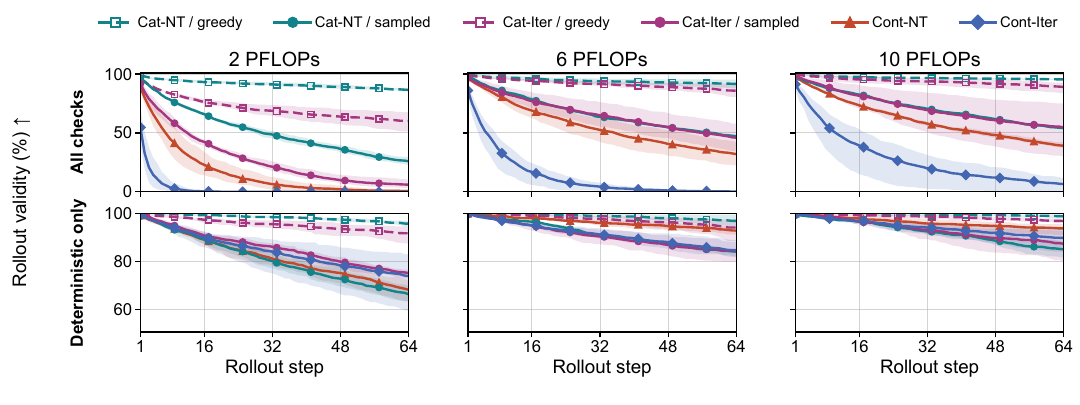}
    \caption{{\textbf{Categorical models maintain higher rollout validity, with the largest differences arising from stochastic dynamics.} Curves show the percentage of rollouts remaining valid through each horizon, with all checks (top) and deterministic only check (bottom) across training compute.}}
    \label{fig:strict_long_horizon}
\end{figure}

\noindent \textbf{Finding 2: Continuous models struggle more to learn stochastic elements at long {horizons}.}
Next, we evaluate models' behaviors at long {horizons}, through rollouts of 64 steps. For each rollout step, we verify the validity of generated transitions against the expected environment behavior. We report rollout validity (Figure \ref{fig:strict_long_horizon}), i.e. how many trajectories have all valid transitions at each rollout step.
We report validity on deterministic elements only, as well as all elements (including the stochastic ghosts). {Both evaluations use the same rollouts; the deterministic-component evaluation excludes ghost-specific constraints.} Detailed results are available in Appendix \ref{sec:supp:detailed_xp}.
Our results show that models achieve similar performance for deterministic transitions, with categorical models (greedy inference) outperforming models with stochastic inference.
We observe substantially more variability for the stochastic elements, with categorical models consistently achieving better performance at all compute budgets. Greedy inference again achieves the best overall results. \ContNT{} consistently improves throughout training, slowly closing the gap with categorical models, although not matching their performance.
The two categorical models approach similar rollout validity at the largest budget, with \CatNT{} achieving stronger results at lower compute.
Lastly \ContIter{} appears to struggle the most to learn the ghost mechanics, with very poor initial performance, slow improvements over compute budgets and a substantial gap remaining {at the largest training budget considered}. {This} experiment suggests that continuous models struggle to model stochastic dynamics.
This aligns with qualitative observations that generative world models can struggle to preserve visually and dynamically uncertain entities over time,
including blurry or temporally inconsistent opponent predictions \citep{DIAMOND,Genie2}.

\noindent \textbf{Finding 3: Categorical models demonstrate better {training-compute} efficiency {for rollout validity}.}
As shown in Figure \ref{fig:strict_long_horizon},
categorical models {achieve higher rollout validity} with smaller {training-compute budgets}.
With greedy decoding, \CatNT{} retains approximately 86\% rollout validity
at step 64 after only 2 PFLOPs of training.
Comparing models on deterministic {components} only shows much closer {performance, indicating that the observed compute gap is largely associated with stochastic dynamics}.

\noindent \textbf{Finding 4: Transition validity does not imply calibrated stochastic predictions, sampled NT type models achieve best calibration.} The validity metrics reported above measure whether generated transitions obey the environment's rules, but not whether different valid outcomes occur with the correct probabilities. 
To understand how these relate, we assess calibration using total variation (TV) between the empirical distribution of predicted next-step ghost configurations and {the simulator's exact conditional distribution over those configurations}.
{Figure~\ref{fig:next_frame}(c) shows that sampled \CatNT{} achieves the lowest empirical TV at every evaluated compute budget. At 10 PFLOPs, mean TV is $0.30$ for sampled \CatNT{} and $0.67$ for sampled \CatIter{}. Both continuous models improve substantially between 2 and 10 PFLOPs, reaching $0.33$ for \ContNT{} and $0.34$ for \ContIter{}. They therefore outperform sampled \CatIter{} in calibration at the largest budget, despite their weaker validity.}
This reversal in model ranking highlights the need to evaluate calibration alongside validity.
The results also suggest an important direction for future work on \CatIter{}-type models specifically to further understand and improve the potential trade-offs between validity, calibration, and inference speed {(evaluation details in Appendix~\ref{app:pacman-metrics})}.

\begin{figure}[t]
     \centering
 
     \includegraphics[width=0.96\linewidth]
         {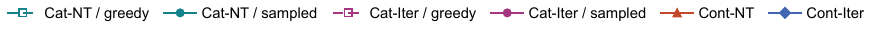}
 
     \begin{subfigure}[t]{0.34\linewidth}
         \centering
         \phantomsubcaption
         \begin{overpic}[width=\linewidth]
             {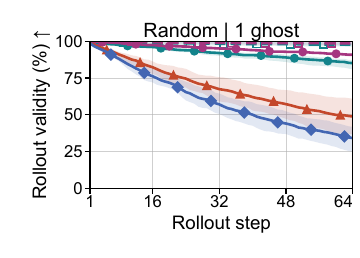}
         \end{overpic}
     \end{subfigure}
     \hfill
     \begin{subfigure}[t]{0.283\linewidth}
         \centering
         \phantomsubcaption
         \begin{overpic}[width=\linewidth]
             {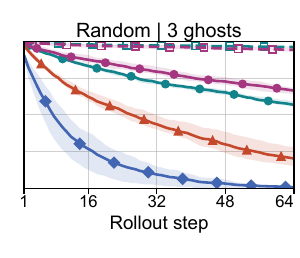}
         \end{overpic}
     \end{subfigure}
     \hfill
     \begin{subfigure}[t]{0.283\linewidth}
         \centering
         \phantomsubcaption
         \begin{overpic}[width=\linewidth]
             {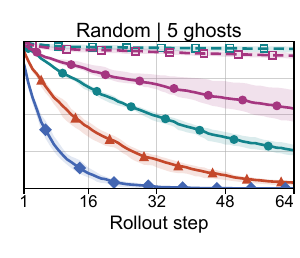}
         \end{overpic}
     \end{subfigure}
 
     \vspace{-0.4em}
 
     \begin{subfigure}[t]{0.34\linewidth}
         \centering
         \phantomsubcaption
         \begin{overpic}[width=\linewidth]
             {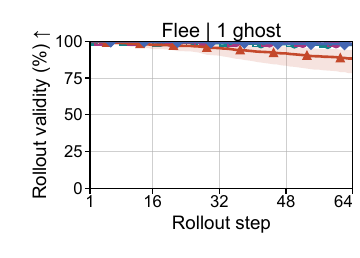}
         \end{overpic}
     \end{subfigure}
     \hfill
     \begin{subfigure}[t]{0.283\linewidth}
         \centering
         \phantomsubcaption
         \begin{overpic}[width=\linewidth]
             {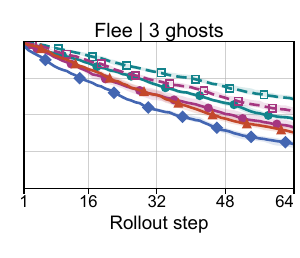}
         \end{overpic}
     \end{subfigure}
     \hfill
     \begin{subfigure}[t]{0.283\linewidth}
         \centering
         \phantomsubcaption
         \begin{overpic}[width=\linewidth]
             {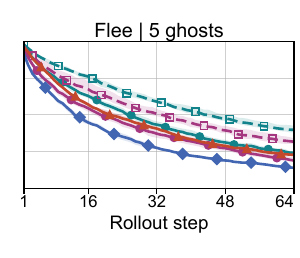}
         \end{overpic}
     \end{subfigure}
 
     \caption{{
         \textbf{Greedy categorical decoding is most robust to increasing complexity.}
         Curves show rollout validity with one, three, or five ghosts under Random (top)
         and Flee dynamics (bottom).
     }}
     \label{fig:complexity-ablation}
 \end{figure}

\noindent \textbf{{Finding 5: Greedy categorical models are more robust to increasing ghost count.}}
We examine how the models' %
rollout validity changes as the ghost dynamics become more demanding. We vary the number of ghosts (one, three, or five) and their behavior (Random or Flee), while keeping the remaining environment settings and training horizon fixed. %
{Figure~\ref{fig:complexity-ablation} shows that categorical models achieve higher rollout validity across most conditions. Greedy \CatNT{} is the most robust approach overall, while greedy \CatIter{} also performs strongly, although its validity decreases as the number of Flee ghosts increases.}
{Greedy decoding improves rollout validity for both categorical models.}
{Sampled decoding remains less reliable: with five Random ghosts, sampled \CatIter{} retains approximately 50\% rollout validity at step 64.}
{Increasing ghost count reduces mean rollout validity. The decline is larger for continuous models than for greedily decoded categorical models, particularly under Random dynamics; sampled categorical models also degrade substantially.}
{Both continuous models achieve higher mean survival AUC under Flee than Random dynamics at every ghost count. Categorical models favor Random with three or five ghosts, but Flee with one (Table~\ref{tab:pacman-overall-strict-auc}).}
Appendix~\ref{app:pacman-structural-semantic} further examines structural failures and adherence to the ghost policy in Flee rollouts.

\subsection{``3D Maze'' Controlled Environment}
\label{subsec:env-mazeballs}

\begin{figure*}[t]
  \centering
    \includegraphics[width=\linewidth]{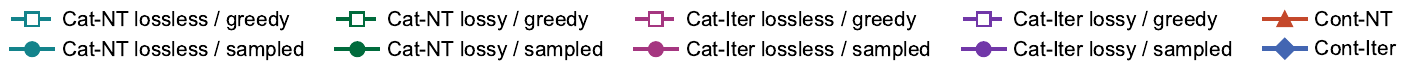}
    \vspace{1pt}

    \begin{subfigure}[t]{0.32\linewidth}
        \centering
        \includegraphics[width=\linewidth]{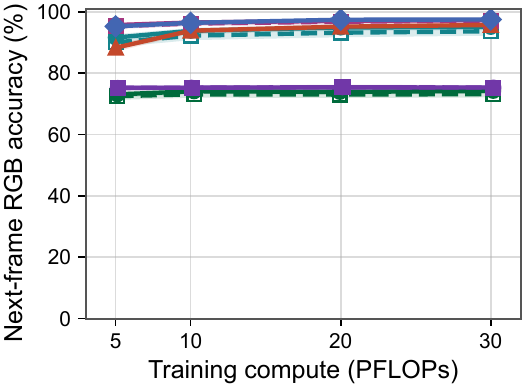}
        \caption{Next-frame RGB accuracy.}
        \label{fig:maze-next-frame}
    \end{subfigure}
    \hfill
    \begin{subfigure}[t]{0.32\linewidth}
        \centering
        \includegraphics[width=\linewidth]{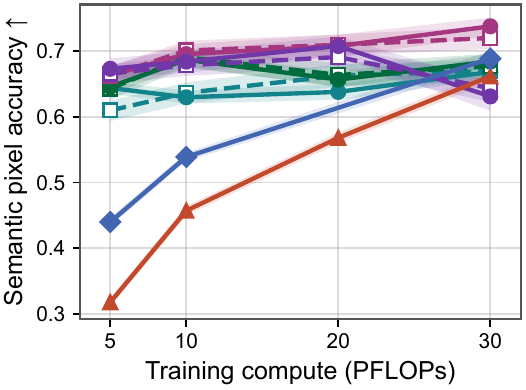}
        \caption{Semantic pixel accuracy.}
        \label{fig:maze-semantic}
    \end{subfigure}
    \hfill
    \begin{subfigure}[t]{0.32\linewidth}
        \centering
        \includegraphics[width=\linewidth]{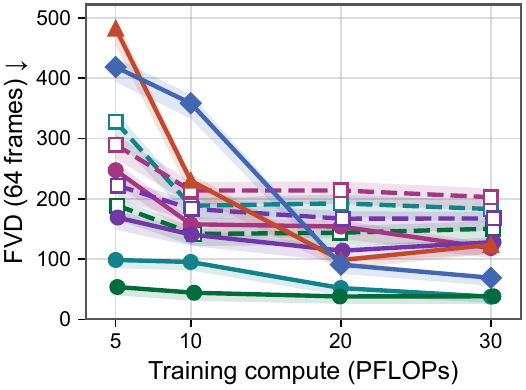}
        \caption{64-frame FVD.}
        \label{fig:maze-fvd}
    \end{subfigure}
  \caption{{\textbf{Lossy tokenization results in worse next-frame RGB accuracy but does not harm semantic or long-horizon metrics.}
  Lossy models have lower exact next-frame RGB accuracy in (a), but remain competitive on
  (b) mean semantic pixel accuracy over 64-step rollouts and (c) 64-frame FVD.
  Here, shading shows pointwise 95\% bootstrap intervals over evaluation cases.}}
  \label{fig:maze-compute}
\end{figure*}

In addition to the 2D pixel-navigation environment, we created a partially observable 3D environment,
which renders the same grid-based map in 3D from a first-person perspective, with continuous, rather than discrete navigation.
{Non-static} elements are included in the form of three bouncing spheres, randomly positioned at the start of the trajectory, with random initial velocities. These bounce off walls but do not interact with the agent. Sphere dynamics are deterministic given their positions and velocities, but these may be uncertain from the observed history.
To prevent visual complexity from obscuring other signals, the rendering uses only solid colors, with no textures or shading. 
Spheres, walls, floor, and ceiling are distinguished by color.
Training trajectories are generated by traversing random routes through the maze, resetting the environment when a randomly chosen goal is reached. See a visual example in Figure \ref{fig:maze} and additional details in Appendix \ref{app:maze-details}.

\noindent \textbf{Image encoding.}
We consider three representations of the environment: discrete lossless, discrete lossy, mirroring common categorical modelling approaches, and continuous. We begin by rendering each frame as a $64\times64\times3$ RGB image and divide this frame into $64\times1\times3$ vertical strips. Continuous models directly operate with these strips as inputs. For the lossless discrete representation, we tokenize each strip using a lookup table. 
The simple rendering allows a vocabulary of 10,000 entries to cover all column patterns encountered in the training and evaluation data.
For the lossy representation, we use a small VQ-VAE {\citep{vqvae}} whose encoder comprises two convolutional layers with $8 \times 3$ kernels and stride $(8,1)$. This produces 64 discrete tokens per frame, one per column, while incorporating neighboring-column context.

\noindent \textbf{Evaluation metrics.} {The exact dynamics-validity analysis used in Pac-Man is not feasible even in this controlled environment, given its partial observability and continuous actions.}
{We therefore use the following three metrics:} \textbf{{Next-frame RGB accuracy}} 
measures exact RGB agreement with the reference next frame.
\textbf{Fr\'echet Video Distance (FVD)} \citep{FVD} is measured over long horizon rollouts to evaluate distributional alignment with ground truth rollouts. \textbf{{Semantic pixel accuracy}} measures pixel-wise agreement with the reference on three classes: wall, spheres, and background. Pixels are classified using the environment\textquoteright{}s color palette.

\subsubsection{Findings in 3D Maze}
\label{subsec:maze-findings}
\noindent \textbf{Finding 1: Lossy tokenization {limits next-frame RGB accuracy}, while lossless and continuous models reach similarly high accuracy.} As in our first Pixel Pac-Man experiment, we measure our models' ability to generate the next frame in a sequence correctly. Figure \ref{fig:maze-next-frame} compares next frame RGB accuracy over training budgets. {Over the evaluated training budgets, lossy models remain below the lossless and continuous models, which improve to similarly near-perfect accuracy.}

\noindent \textbf{Finding 2: The visual accuracy gap between lossy and lossless categorical models does not translate into a consistent disadvantage in long-horizon metrics.}
Next, we evaluate how models behave over long horizons, computing both FVD (Figure \ref{fig:maze-fvd}) and semantic accuracy (Figure \ref{fig:maze-semantic}) over 64-step rollouts. 
Lossy models remain competitive with their lossless counterparts, indicating that lower next-frame accuracy does not consistently translate into worse semantic accuracy or FVD.
We note that compression may have a greater impact on dynamics that depend on fine visual details. Nevertheless, this result highlights the importance of disentangling visual quality from temporal modeling performance.

\noindent \textbf{Finding 3: Categorical models achieve stronger long-horizon metrics at low training budgets, while continuous models narrow the gap with additional compute.}
{As in Pixel Pac-Man, categorical models achieve stronger rollout metrics at lower training compute.}
Continuous models substantially narrow the gap by 30 PFLOPs, although differences between models and decoding procedures remain.
Unlike the stochastic ghost dynamics in Pixel Pac-Man, uncertainty in the maze arises from partial observability, including spheres entering view with previously unobserved positions or velocities.
In addition, while Pixel Pac-Man validity results heavily favored greedy inference, sampling substantially improves categorical FVD with respect to greedy decoding in this maze setting,
without a correspondingly large change in other metrics.

\section{Conclusion}

We presented a controlled comparison of categorical and continuous visual world models across 2D and 3D environments that were specifically constructed for systematic analysis. {In a lossless 2D environment, we found that categorical models achieved higher rollout validity at lower training compute, particularly for stochastic dynamics.}
We also empirically demonstrated that models achieving high validity can fail to be calibrated, with (non-deterministic) next token prediction (NT) models {achieving better calibration than} iterative models that exemplify today's predominant diffusion-type models.
In a 3D maze environment, we found that differences between lossless and lossy tokenization were outweighed by differences due to model type, with categorical (token-based) models outperforming models with continuous representations in semantic pixel accuracy and FVD.
Findings in both settings challenge today's mainstream large-scale world model training efforts and suggest a closer look at {categorical prediction for learning valid dynamics efficiently}.

\bibliography{iclr2027_conference}
\bibliographystyle{iclr2027_conference}

\appendix
\section*{Appendix} \addcontentsline{toc}{section}{Appendix}

\section{Implementation and Training Details}
\label{app:world_model_training}

Table~\ref{tab:shared_training_details} provides training details that are shared across architectures.
{We estimate training compute using the approximation $6N_{\mathrm{train}}D$, where $N_{\mathrm{train}}$ is the number of trainable world-model parameters and $D$ counts observation and action tokens processed during training. This estimates computed for the backbone forward and backward passes. Encoder training is excluded.}

\begin{table}[ht]
\centering
\footnotesize
\caption{
Shared training settings. {Compute budgets refer to the maximum budgets for the main comparisons.}
}
\label{tab:shared_training_details}
\begin{tabularx}{\columnwidth}{@{}lXX@{}}
\toprule
\textbf{Setting} & \textbf{{Pixel Pac-Man}} & \textbf{3D Maze} \\
\midrule
Trajectories per epoch & 2,048 & 2,048 \\
Frames per trajectory & 16 & 16 \\
Batch size            & 128 & 128 \\
Training budget       & {10 PFLOPs} & 30 PFLOPs \\
Training regime       & Teacher forcing & Teacher forcing \\
Optimizer             & AdamW & AdamW \\
Learning rate         & Constant $4\times10^{-4}$
                      & Constant $4\times10^{-4}$ \\
Weight decay          & 0 & 0 \\
Gradient clipping     & Global norm 1.0 & Global norm 1.0 \\
Transformer depth     & 6 blocks & 6 blocks \\
Attention heads       & 4 & 4 \\
MLP ratio             & 4 & 4 \\
nanoGPT width         & 44 & 44 \\
DiT width             & 40 & 36 \\
Continuous encoder    & Linear projection
                      & 2-layer 1D CNN, 48 channels,
                        then linear projection \\
Validation trajectories & 128 & 512 \\
Training seeds        & 3 & 1  \\
\bottomrule
\end{tabularx}
\end{table}

Table~\ref{tab:model_specific_training_details} details {architecture-specific} design choices.

\begin{table*}[ht]
\centering
\footnotesize
\caption{Model-specific training and inference settings for the main
comparisons.}
\label{tab:model_specific_training_details}
\begin{tabularx}{\textwidth}{@{}lXX@{}}
\toprule
\textbf{Model} & \textbf{Training details} & \textbf{Inference details} \\
\midrule

\CatNT{}
& Cross-entropy on the final target frame; visual input embeddings
are tied to the output classifier.
& Sequential patch generation with greedy decoding ($T=0$) or
categorical sampling ($T=1$), without top-$k$ truncation. \\

\ContNT{}
& Conditional flow matching with three residual MLP blocks of hidden
width 23; logit-normal time sampling with parameters $\mu=0$,
$\sigma=1$, over 1,000 discrete timesteps; training flow shift 3.
& Each patch is initialized from Gaussian noise and generated
separately before being fed back into the causal sequence.
Pac-Man: 50 Euler evaluations per patch, inference shift 1.
3D maze: 20 Euler evaluations per patch, inference shift 3. \\

\CatIter{}
& The number of masked target patches is sampled uniformly from
$\{0,\ldots,P\}$, where $P$ is the number of patches per image; cross-entropy is evaluated only at masked positions.
At least one masked position is enforced per training batch.
& $16$ reveal iterations for Pac-Man and $20$ for the 3D maze,
using a linear schedule. Remaining
tokens are predicted using greedy decoding ($T=0$) or categorical
sampling ($T=1$); the highest-confidence predictions are
permanently committed at each iteration. \\

\ContIter{}
& Whole-frame conditional flow matching with logit-normal time
sampling with parameters $\mu=0$, $\sigma=1$, over 1,000 discrete
timesteps; training flow shift 3.
& The complete frame is initialized from Gaussian noise and updated
jointly. Pac-Man: 50 Euler evaluations per frame.
3D maze: 20 Euler evaluations per frame.
Inference shift 3 in both environments. \\

\bottomrule
\end{tabularx}
\end{table*}

{For the ghost-dynamics ablation (Figures~\ref{fig:complexity-ablation} and~\ref{fig:pacman-structural-semantic}, and Table~\ref{tab:pacman-overall-strict-auc}), we use the saved checkpoint nearest to $12$ PFLOPs per run and $20$ flow-matching inference steps for continuous models.}

Table \ref{tab:vqvae_training} provides implementation details for the VQ-VAE trained on the 3D maze dataset. 

\begin{table}[ht]
\centering
\caption{Column VQ-VAE architecture and training configuration.}
\label{tab:vqvae_training}
\begin{tabular}{ll}
\hline
Setting & Value \\
\hline
Input & $64 \times 64 \times 3$ RGB, scaled to $[0,1]$ \\
Tokens per frame & 64 ($1 \times 64$ grid) \\
Codebook size & 256 \\
Embedding dimension & 32 \\
Hidden channels & 64 \\
Encoder & Two convolutions, ReLU between layers \\
Encoder kernels / strides & $8 \times 3$ / $(8,1)$ in both layers \\
Decoder & Two transposed convolutions \\
Decoder residual blocks & 2, before upsampling \\
Reconstruction loss & Mean absolute error ($L_1$) \\
Codebook loss & Mean squared error \\
Commitment coefficient & $0.25$ \\
Optimiser & Adam \\
Learning rate & $10^{-3}$, constant \\
Gradient clipping & Global norm of 1.0 \\
Training steps & 300,000 \\
Batch size & 64 \\
Data sampling & Shuffled frames from 8 fresh trajectories per buffer \\
\hline
\end{tabular}
\end{table}

\begin{samepage}
\subsection{Pac-Man Implementation details}
\label{app:pacman-implementation}

{Training and inference settings are listed in Tables~\ref{tab:shared_training_details} and~\ref{tab:model_specific_training_details}, with the ghost-dynamics exception specified above. The nanoGPT and DiT backbones are approximately matched in parameter count, with additional model-specific capacity in the input embeddings and output heads. Appendix~\ref{app:param_analysis} examines inference-parameter choices.}
\end{samepage}

\subsection{Pac-Man evaluation details}
\label{app:pacman-metrics}

{Evaluation contexts and action sequences are shared across models. Validity results use $512$ contexts per checkpoint and decoding condition. Lines show equal-weight means across training seeds; shading shows their pointwise minimum--maximum range, including in the ghost-dynamics complexity ablation. Calibration uses $32$ next-frame predictions per context on a fixed, model-independent subset of $232$ contexts with exactly identifiable conditional ghost-configuration distributions.}

We evaluate one-step predictions conditioned on 15 observed frames and the next action. \textbf{Transition validity} is the fraction of predicted transitions satisfying the selected simulator constraints, irrespective of earlier failures. For long-horizon evaluation, we generate $H=64$ frames autoregressively under a prescribed action sequence, feeding generated observations back into the model. \textbf{Rollout validity} at horizon $h$ is the fraction of trajectories that remain valid throughout their first $h$ generated transitions. Once a trajectory fails, subsequent valid transitions do not restore its rollout validity. The first generated transition is indexed by $t=1$, as conditioning frames are excluded from all reported metrics.

\paragraph{State representation and validity criteria.}
All predictions are evaluated in a common representation with wall, pellet, Pac-Man, and ghost occupancy channels. Categorical tokens decode into patches of allowed cell states. Continuous RGB predictions are mapped to the nearest allowed palette colour at each cell; RGB-native rollouts also apply this projection before feeding predictions back into the model. Consequently, cell-state validity is enforced by construction. This does not enforce board-level or temporal consistency, which require the checks below.

\paragraph{Component checks.}
Pac-Man validity requires exactly one Pac-Man and movement consistent with the supplied action and wall layout. Ghost validity checks occupancy and movement permitted by the configured ghost policy. For stochastic ghosts, any permitted movement is accepted, regardless of whether it matches the movement in a reference trajectory. Board validity requires wall preservation and pellet consistency jointly. Pellet checks distinguish changes in visibility from changes in underlying presence: ghosts may hide pellets without consuming them, and revealing a previously hidden pellet is not counted as
creation.

\paragraph{History consistency.}
A sequence of locally valid transitions need not admit a consistent hidden-state history. The full rollout validator therefore propagates compatible hidden-state hypotheses from the observed context through the generated sequence, including hypotheses about occluded pellets. Local transition checks do not require the earlier generated prefix to have remained valid.

\paragraph{Episode boundaries.}
Evaluation continues across episodes. An observed reset crossing is checked against the simulator's clean-reset rule, including the prescribed agent positions and restored pellet layout, accounting for occlusion. The action supplied at the reset boundary is ignored, as in the simulator. Capture and episode termination are not themselves validity failures. A terminal observation at the final forecast step provides no subsequent reset transition to evaluate.

\paragraph{Ghost-excluded evaluation.}
We score the same generated trajectories with and without ghost-specific constraints. {This comparison distinguishes validity on deterministic components (Pac-Man and board) from validity that also accounts for stochastic ghost dynamics.} Ghost-excluded evaluation omits ghost checks and joint reset reachability, while retaining local Pac-Man and board checks, fresh-pellet requirements at resets, and collision rejection at definite reset crossings.

\subsection{Maze environment evaluation details}
\label{app:maze-evaluation-details}
{Training and inference settings are listed in Tables~\ref{tab:shared_training_details} and~\ref{tab:model_specific_training_details}.}
{Uncertainty is quantified using pointwise 95\% bootstrap intervals ($2{,}000$ resamples; $n=512$ evaluation cases).}

\section{Additional maze environment details}
\label{app:maze-details}

We provide additional detail on the maze environment design and data construction. 

The agent navigates via two continuous joystick values \textendash{} a steering value (where -1 means turn full-speed left, 1 means turn full-speed right, and 0 is a turn no-op), and a forwards motion value (where 1 means full-speed forwards, and 0 means stand still). Agents cannot walk through walls, but can slide along them, so a wall collision does not necessarily leave the agent stationary.

Camera pitch is fixed so that the agent is always looking straight ahead in the direction of movement, and wall height and ball elevation are uniform, such that the geometry of the scene is always symmetrical about the horizontal axis. Internal walls are all one color, while the North, South, East and West external walls each have their own color. Spheres are drawn in solid yellow, the floor in solid green, and the ceiling in solid red. 

\section{Detailed experiment results}
\label{sec:supp:detailed_xp}

{This section provides additional Pixel Pac-Man results.}

\begin{figure}[!htbp]
    \centering
    \includegraphics[width=\linewidth]{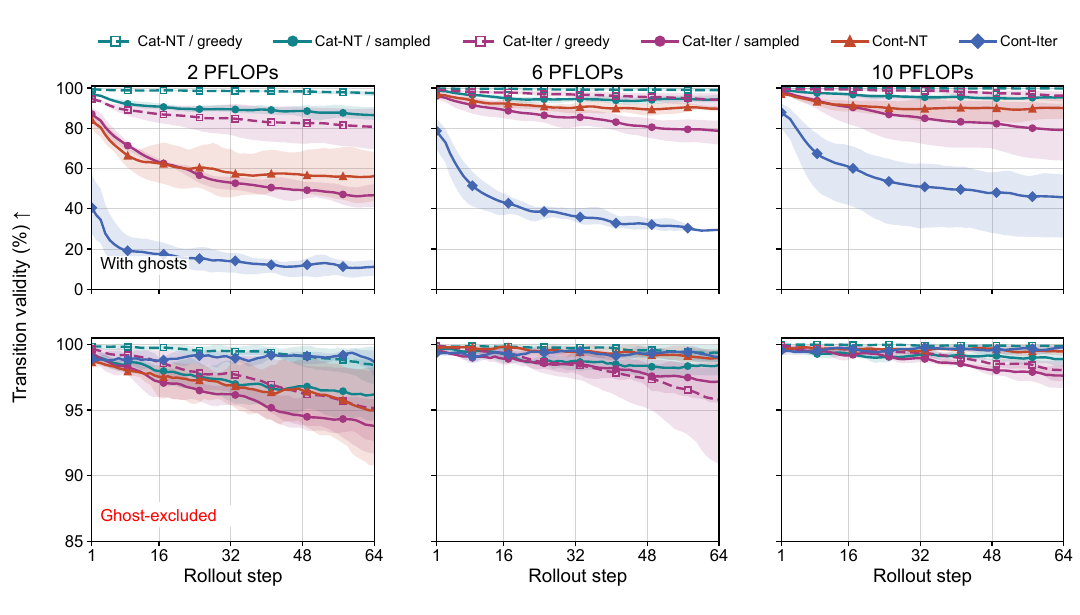}
    \caption{{\textbf{Continuous models struggle with stochastic dynamics throughout rollout, particularly \ContIter{}.} Each transition is evaluated independently of earlier failures. Validity on deterministic components remains high (bottom), whereas including stochastic ghost dynamics reveals persistent errors (top).}}
    \label{fig:local_long_horizon}
\end{figure}

First, we complement the strict rollout-validity results in Figure~\ref{fig:strict_long_horizon} with local transition validity in Figure~\ref{fig:local_long_horizon}. This metric evaluates each predicted transition independently of earlier errors, showing whether a model returns to producing locally valid transitions after a failure; such local recovery does not restore the validity of the preceding rollout history. As before, we compare evaluation with all checks against {deterministic-component} evaluation, which excludes ghost-dynamics checks while retaining Pac-Man, board, collision, and reset constraints. {Transition validity on deterministic components} remains high for all approaches, whereas including {stochastic} ghost dynamics reveals persistent local errors, particularly for the continuous models and most strongly for \ContIter{}. Thus, the decay in strict rollout validity observed in Figure~\ref{fig:strict_long_horizon} is not solely an artefact of permanently recording an isolated early failure: the continuous models continue to produce locally invalid transitions throughout the rollout.

\begin{figure}[!htbp]
    \centering
    \includegraphics[width=\linewidth]{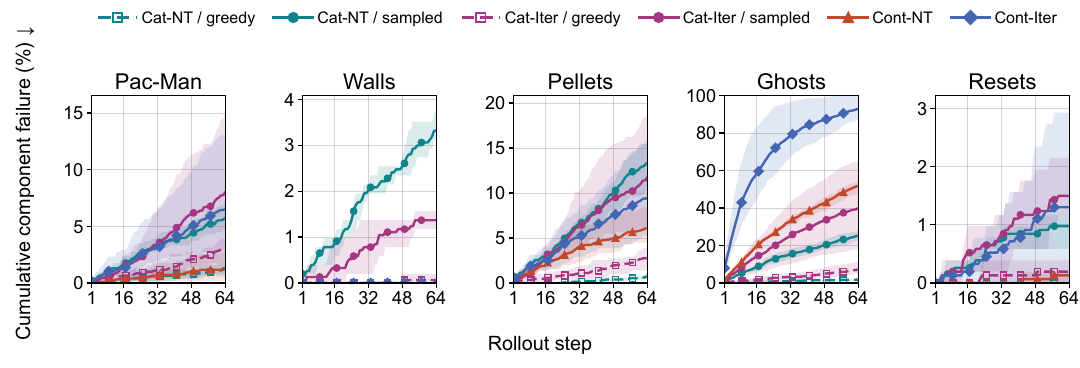}
    \caption{{\textbf{Stochastic ghost dynamics are the largest source of rollout failure, particularly for \ContIter{}.} Panels show the percentage of rollouts with at least one violation of each component by horizon. A rollout may contribute to multiple panels. Note the different vertical scales.}}
    \label{fig:component_failure}
\end{figure}

{Figure~\ref{fig:component_failure} breaks down cumulative failures by component at approximately $10$ PFLOPs. Deterministic-component failures are generally smaller than ghost failures, although sampled categorical models also accumulate Pac-Man and pellet errors.}
Wall failures are concentrated in sampled categorical models, with only
negligible failure rates under greedy decoding.

\FloatBarrier
\raggedbottom
\begingroup
\setlength{\intextsep}{6pt plus 2pt minus 2pt}
\subsection{Parameter Analysis Experiments}
\label{app:param_analysis}

\noindent \textbf{Continuous model inference parameter study.} {We evaluate continuous-model rollouts with $20$, $50$, and $100$ inference steps across all training seeds.}
Figure~\ref{fig:supp:cont_inference} reports transition validity, strict rollout validity, and mean
valid-prefix length. These measure, respectively, the percentage of valid
transitions, the percentage of rollouts remaining valid through all
64 generated frames, and the average number of consecutive valid frames
before the first failure.
{Increasing the step count from $20$ to $50$ improves rollout quality for both continuous models. Beyond $50$ steps, \ContIter{} appears to plateau, while \ContNT{} improves more modestly. These gains motivate the main-comparison settings in Table~\ref{tab:model_specific_training_details}.}

\begin{figure}[H]
    \centering
    \includegraphics[width=\linewidth,trim=0 12bp 0 5bp,clip]{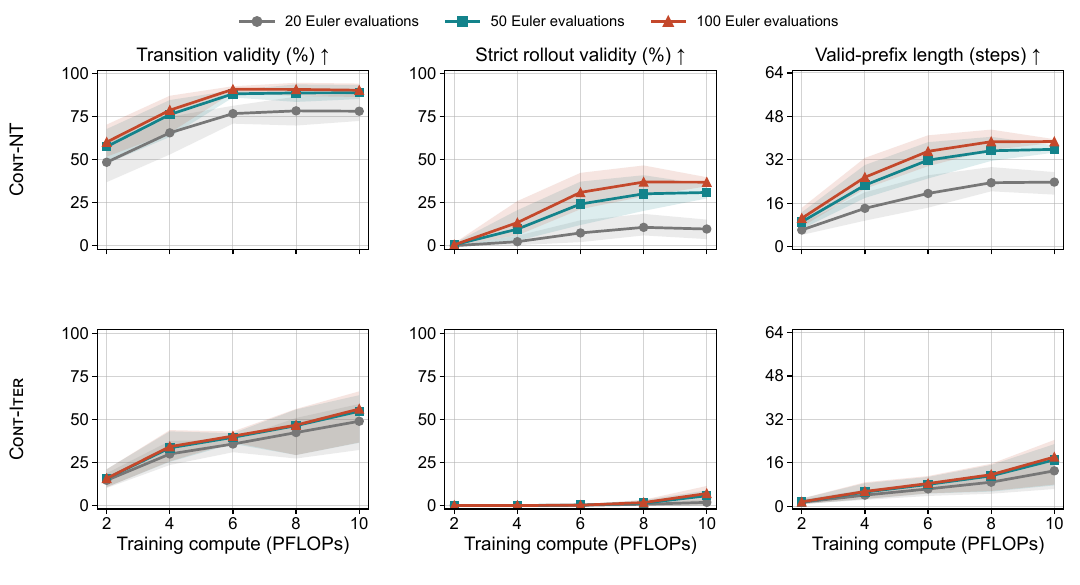}
    \caption{{\textbf{Increasing inference steps improves continuous-model rollouts, with smaller gains beyond $50$ steps.} Rows show \ContNT{} (top) and \ContIter{} (bottom); columns show transition validity, rollout validity through $64$ steps, and mean valid-prefix length.}}
    \label{fig:supp:cont_inference}
\end{figure}

\begin{figure}[H]
    \centering
    \includegraphics[width=\linewidth,trim=0 12bp 0 5bp,clip]{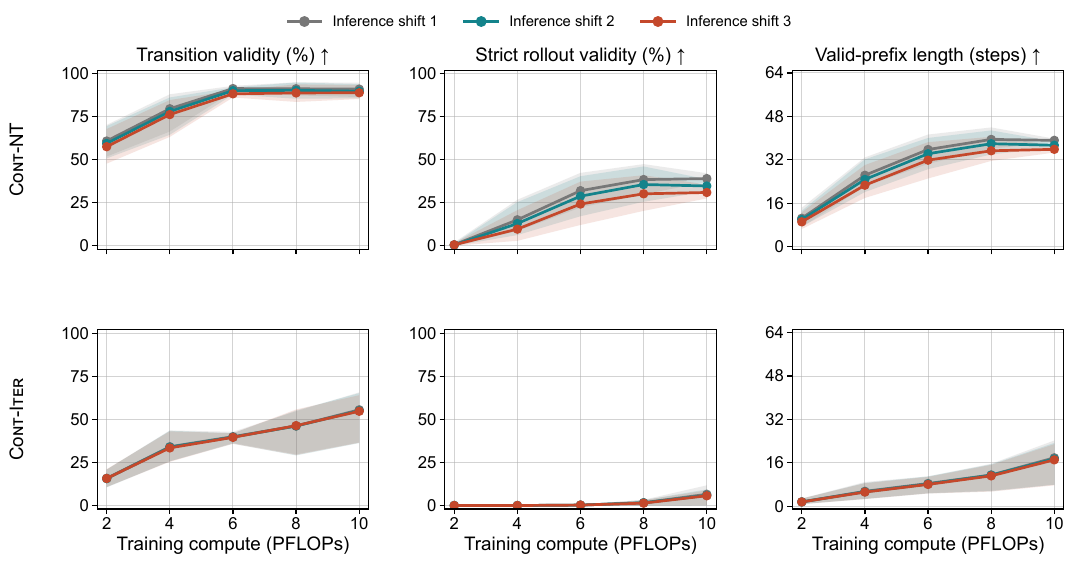}
    \caption{{\textbf{\ContNT{} favors smaller flow shifts, while \ContIter{} is largely insensitive to this parameter.} Rows show \ContNT{} (top) and \ContIter{} (bottom); columns show transition validity, rollout validity through $64$ steps, and mean valid-prefix length.}}
    \label{fig:supp:flow_shift}
\end{figure}

{We also vary the inference flow shift (Figure~\ref{fig:supp:flow_shift}), which controls where steps are placed along the noise-to-target trajectory. Higher values allocate more steps to the early, noisier part. Flow shift has little effect on \ContIter{}, whereas \ContNT{} favors smaller values. Overall, changing the flow shift has less effect than changing the number of inference steps.}

\noindent \textbf{Cat-Iter model inference parameter study.}
{For \CatIter{}, we test $4$, $8$, $10$, $12$, and $16$ inference steps at temperatures $T\in\{0,0.5,1,1.5\}$ (Figure~\ref{fig:supp:maskgit_inference}). Among these settings, $16$ steps with greedy decoding achieve the best performance. Greedy decoding improves results, but the choice of step count still has a substantial effect.}

\begin{figure}[H]
    \centering
    \includegraphics[width=\linewidth]{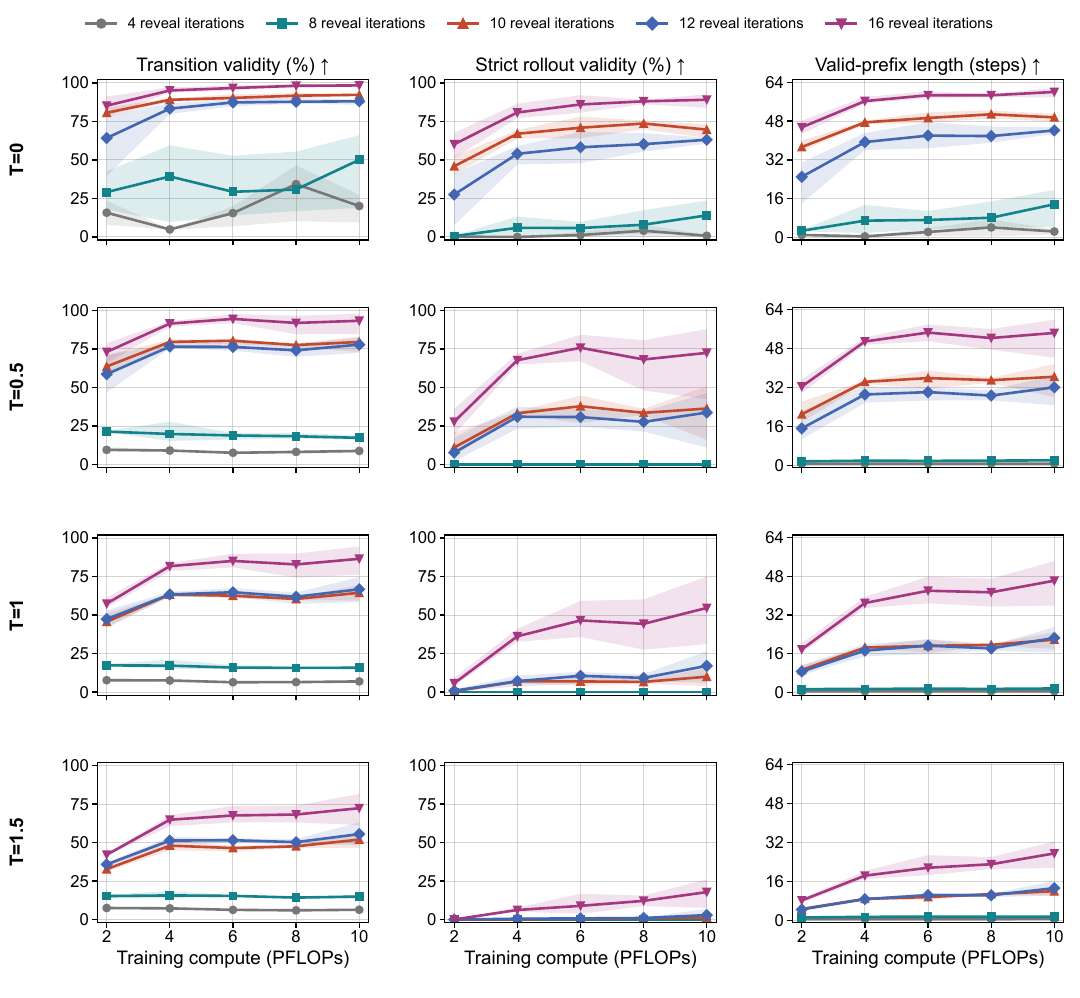}
    \caption{{\textbf{Sixteen inference steps with greedy decoding give the strongest \CatIter{} results.} Step count remains important even under greedy decoding. Rows vary sampling temperature; columns show transition validity, rollout validity through $64$ steps, and mean valid-prefix length.}}
    \label{fig:supp:maskgit_inference}
\end{figure}
\FloatBarrier
\endgroup

\subsection{Structural and Semantic Validity Under Different Ghost Dynamics}
\label{app:pacman-structural-semantic}

\begin{figure}[!htbp]
     \centering
 
     \includegraphics[width=0.96\linewidth]
         {images/three_seed_matched/bootstrap_shared_legend.pdf}
     \vspace{-0.4em}
 
     \begin{subfigure}[t]{0.34\linewidth}
         \centering
         \phantomsubcaption
         \begin{overpic}[width=\linewidth]
             {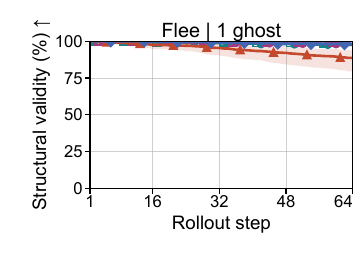}
             \put(27,25){\colorbox{white}{\textbf{(\thesubfigure)}}}
         \end{overpic}
     \end{subfigure}
     \hfill
     \begin{subfigure}[t]{0.283\linewidth}
         \centering
         \phantomsubcaption
         \begin{overpic}[width=\linewidth]
             {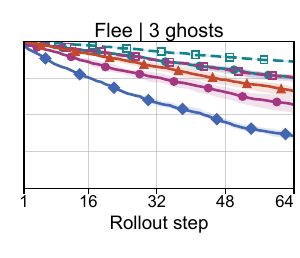}
             \put(12,30){\colorbox{white}{\textbf{(\thesubfigure)}}}
         \end{overpic}
     \end{subfigure}
     \hfill
     \begin{subfigure}[t]{0.283\linewidth}
         \centering
         \phantomsubcaption
         \begin{overpic}[width=\linewidth]
             {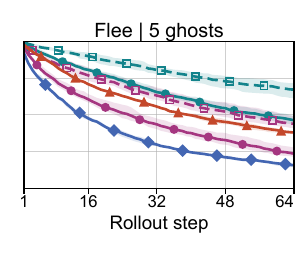}
             \put(12,30){\colorbox{white}{\textbf{(\thesubfigure)}}}
         \end{overpic}
     \end{subfigure}
 
     \vspace{-0.4em}
 
     \begin{subfigure}[t]{0.34\linewidth}
         \centering
         \phantomsubcaption
         \begin{overpic}[width=\linewidth]
             {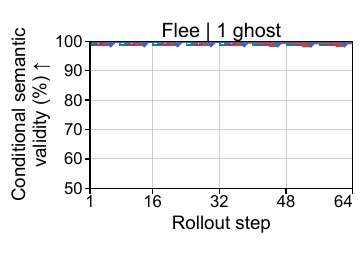}
             \put(27,25){\colorbox{white}{\textbf{(\thesubfigure)}}}
         \end{overpic}
     \end{subfigure}
     \hfill
     \begin{subfigure}[t]{0.283\linewidth}
         \centering
         \phantomsubcaption
         \begin{overpic}[width=\linewidth]
             {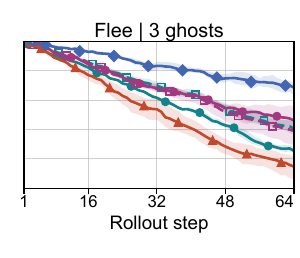}
             \put(12,30){\colorbox{white}{\textbf{(\thesubfigure)}}}
         \end{overpic}
     \end{subfigure}
     \hfill
     \begin{subfigure}[t]{0.283\linewidth}
         \centering
         \phantomsubcaption
         \begin{overpic}[width=\linewidth]
             {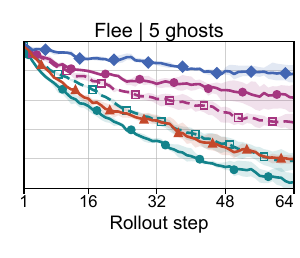}
             \put(12,30){\colorbox{white}{\textbf{(\thesubfigure)}}}
         \end{overpic}
     \end{subfigure}
 
     \caption{{\textbf{Structural failures explain much, but not all, of the continuous-model deficit under Flee dynamics.}
    Top: structural validity, allowing any legal ghost direction while retaining all other constraints.
    Bottom: fleeing-policy adherence among structurally valid prefixes.
    Columns show one, three, and five ghosts.
    Lines show means across training seeds; shading shows their pointwise minimum--maximum range.}}
     \label{fig:pacman-structural-semantic}
 \end{figure}

{Figure~\ref{fig:complexity-ablation} and Table~\ref{tab:pacman-overall-strict-auc} show that the Random/Flee ordering depends on ghost count and model family. All models have higher mean survival AUC under Flee with one ghost. With three or five ghosts, continuous models favor Flee, whereas categorical models favor Random. To examine the sources of failure, Figure~\ref{fig:pacman-structural-semantic} separates structural consistency from adherence to the fleeing policy.}

\begin{table}[H]
\centering
\caption{{Greedy categorical decoding achieves the highest mean rollout-validity AUC in each setting. AUC is the mean rollout-validity percentage over steps 1--64 (higher is better), averaged across training seeds.}}
\label{tab:pacman-overall-strict-auc}
\begin{tabular}{@{}lrrrrrr@{}}
 \toprule
 Method & Random (1) & Random (3) & Random (5) & Flee (1) & Flee (3) & Flee (5) \\
 \midrule
 {\CatNT{} ($T=0$)} & 98.45 & 97.34 & 95.80 & 99.24 & 79.35 & 61.51 \\
 {\CatNT{} ($T=1$)} & 92.15 & 73.89 & 52.62 & 98.74 & 71.11 & 47.89 \\
 {\CatIter{} ($T=0$)} & 99.09 & 96.25 & 92.95 & 99.52 & 74.06 & 53.97 \\
 {\CatIter{} ($T=1$)} & 95.16 & 79.97 & 71.40 & 99.04 & 65.39 & 42.08 \\
 {\ContNT{}} & 69.63 & 47.40 & 27.53 & 94.60 & 64.52 & 44.98 \\
 {\ContIter{}} & 60.20 & 18.61 & 9.82 & 99.16 & 56.47 & 34.15 \\
 \bottomrule
\end{tabular}
\end{table}

We measure \emph{structural validity} by relaxing the ghost-policy constraint: ghosts may take any legal direction rather than following the fleeing preference and tie-break. All other checks remain active, including walls, ghost occupancy and count, legal movement, {Pac-Man} dynamics, pellet history, collisions, and resets. A sequence is structurally valid through step $h$ only if its entire prefix satisfies this relaxed evaluator.

Let $S_h$ denote structural validity through step $h$ and $V_h$ denote
validity under the original fleeing rules. Since $V_h\subseteq S_h$,
we report conditional semantic validity as
\[
100\,\Pr(V_h\mid S_h)
=100\,\frac{N(V_h)}{N(S_h)},
\]
where $N$ counts sequence prefixes satisfying each event.
This is the percentage of structurally valid prefixes that also satisfy
the fleeing policy; it is undefined when no structural survivors remain.
{We compute these ratios separately for each training seed before averaging and taking pointwise extrema.}

\paragraph{Structural failures account for much of the continuous-model deficit.}
 The first row shows substantial losses of structural validity for continuous models as the number of ghosts and rollout length increase. \ContIter{}, in particular, retains high conditional semantic validity despite pronounced structural degradation. \ContNT{} also exhibits substantial policy errors among structural survivors: with three Flee ghosts, its mean structural and conditional semantic validity at step 64 are approximately 66\% and 57\%, respectively. Structural failures therefore account for much of the continuous-model deficit, especially for \ContIter{}, but do not explain it alone. The two rows have different denominators and are not additive error rates.

\paragraph{Possible explanation for the behavior asymmetry.}
One hypothesis is that continuous models predict sharper, more coherent whole frames when the future is closer to unimodal. Conditional on a fully specified state and action, fleeing reduces transition ambiguity relative to random ghost motion. With multiple plausible futures, continuous generation may instead produce less coherent combinations of object positions or appearances, increasing structural failures. A complementary hypothesis is that continuous models can exploit a spatial shortcut, such as learning a smooth ``move away'' vector field, that approximates fleeing dynamics. Random motion offers no equivalent directional regularity. Discrete models may preserve valid frame structure more reliably under stochastic dynamics, while still finding the coordinated fleeing rule difficult. 
These explanations remain speculative:
Figure~\ref{fig:pacman-structural-semantic} evaluates only Flee rollouts
and does not directly measure sharpness or learned vector fields.

\end{document}